\pdfoutput=1
\documentclass{bmvc2k}
\usepackage[latin9]{inputenc}
\usepackage{float}
\usepackage{units}
\usepackage{textcomp}
\usepackage{graphicx}

\makeatletter

\providecommand{\tabularnewline}{\\}

\title{OnionNet: Sharing Features in Cascaded Deep Classifiers}

\addauthor{Martin~Simonovsky}{martin.simonovsky@enpc.fr}{1}
\addauthor{Nikos~Komodakis}{nikos.komodakis@enpc.fr}{1}

\addinstitution{
Imagine Lab\\
Universit\'{e} Paris Est / \'{E}cole des Ponts\\
Paris, France
}

\runninghead{Simonovsky, Komodakis}{OnionNet: Sharing Features in Cascaded ...}

\def\eg{\emph{e.g}\bmvaOneDot}

\def\etal{\emph{et al}\bmvaOneDot}

\def\ie{\emph{i.e}\bmvaOneDot}

\def\wrt{w.r.t\bmvaOneDot}

\makeatother

\begin{document}
\maketitle
\begin{abstract}
The focus of our work is speeding up evaluation of deep neural networks
in retrieval scenarios, where conventional architectures may spend
too much time on negative examples. We propose to replace a monolithic
network with our novel cascade of feature-sharing deep classifiers,
called OnionNet, where subsequent stages may add both new layers as
well as new feature channels to the previous ones. Importantly, intermediate
feature maps are shared among classifiers, preventing them from the
necessity of being recomputed. To accomplish this, the model is trained
end-to-end in a principled way under a joint loss. We validate our
approach in theory and on a synthetic benchmark. As a result demonstrated
in three applications (patch matching, object detection, and image
retrieval), our cascade can operate significantly faster than both
monolithic networks and traditional cascades without sharing at the
cost of marginal decrease in precision.
\end{abstract}

\section{Introduction}

The last several years have seen deep neural networks (DNNs) bringing
tremendous rise in performance to variety of recognition tasks. However,
this often comes at a price of high computational cost at test time,
the reduction of which has recently become a hot topic in deep learning
\cite{faster,Distilling,Zhang15}. Particularly in retrieval scenarios,
large amount of computational time may be spent on negative examples
of varying difficulty. 

A popular remedy is to set up a cascade of multiple classifiers of
increasing strength, called stages \cite{ViolaJones}. Recently, a
pair of independent DNNs was used in a cascade \cite{Angelova15,Zheng15,LongRange}.
Also the Region proposal network of Faster R-CNN \cite{faster} can
be essentially seen as the first stage in a two-stage cascade. While
in the former case both networks receive the raw input and build up
their higher-level representation individually, in the latter case
the stages are finetuned to share their first five convolutional layers.
As these are the most expensive ones to compute \cite{HeSun14}, it
is questionable whether such powerful features are always necessary.

Our observation is that these are the extreme cases of sharing. If
the intermediate representation is not reused, a representation presumably
at least as powerful as before has to be rebuilt in the following
stage and the running time for positive examples suffers. On the other
hand, making the first stage use the representation of the last stage
may lead to losing time on easy negatives.

We address this by proposing OnionNet, a novel architecture where
the next stage extends the feature map set of the previous stage,
preventing repeated computation. Crucially, the architecture is flexible:
the next stage may add both new layers as well as new feature channels,
while reusing the previous ones at the same time. Thus, our stages
do not have to be of increasing depth only, even classifiers of the
same depth but increasing width are still able to share their features.
To accomplish this, the model is trained end-to-end in a principled
way under a joint loss. 

OnionNet is demonstrated in three important tasks: patch matching,
proposal-based object detection, and image retrieval. We achieve substantial
speed-up compared to non-cascaded baselines as well as non-sharing
cascades, with only a marginal loss in precision.

As our main contributions we show that cascaded DNN may offer significant
computational benefits compared to monolithic architectures, propose
a novel cascaded architecture that promotes feature sharing leading
to additional computational advantages, and provide a systematic study
that sheds further light into the time cost behavior of cascaded architectures.

\vspace{-0.01\textheight}

\section{Related Work}

\begin{figure*}
\begin{centering}
\includegraphics[width=0.9\linewidth]{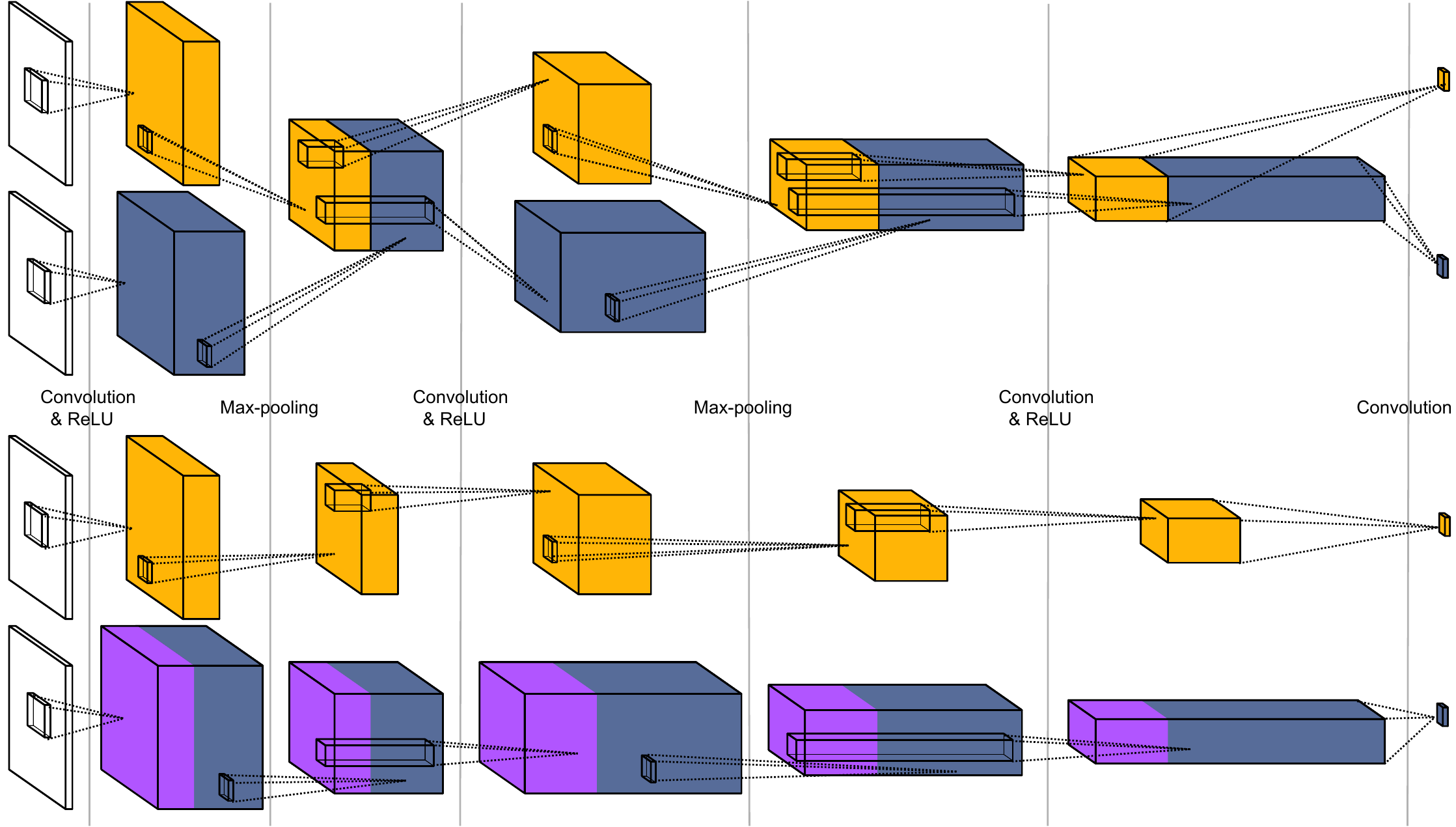}
\par\end{centering}

\caption{\label{fig:method_overview}Feature map sharing. \emph{Top}: Two-stage
OnionNet. \emph{Bottom}: A corresponding non-sharing cascade. In OnionNet,
the first stage (S1, orange) shares its intermediate feature maps
(visualized as cubes) with the second stage (S2, blue). Without sharing
the stages are independent and S2 has to be evaluated fully, recomputing
certain features (purple).}
\end{figure*}

\textbf{Cascades and Sharing.} Whereas in the pioneering work of Viola
and Jones \cite{ViolaJones} stages are distinct and essentially trained
with hard negative mining, the soft cascades of Bourdev and Brandt
\cite{SoftCascade} are trained as a single boosted classifier where
each weak learner has a cumulative score rejection threshold. We are
motivated by the general idea of stages building successively on each
other and realize it in the context of DNNs. Zehnder \etal \cite{Zehnder}
share stages among several class-specific cascades for multi-class
detection, but the stages itself are independent. In deep learning,
the sharing idea of Faster R-CNN \cite{faster} comes probably the
closest to our method. However, their training is less principled
than ours, using a '4-step training algorithm to learn shared features
via alternating optimization'. Moreover, our architecture is more
flexible, as subsequent stages can also add new feature channels besides
new layers.

We also note that the term 'cascades' is overloaded in the literature.
Several authors \cite{Yi13,DeepPose,Dollar2010} speak of cascades
to describe a sequence of stages, evaluated as whole, where one stage
receives the output of another and further refines it. Our cascades
aim for early rejection of negatives.

\textbf{Conditional Execution.} Our approach is also related to conditional
evaluation of networks. In the hierarchical classification with HD-CNN
\cite{hdcnn}, class group specialist networks are executed based
on the prediction of a group classifier, all sharing early layers.
Unlike our approach, HD-CNN aims for precision rather than speed.
Dynamic Capacity Network \cite{DynaCap} uses an entropy-based attention
mechanism to apply a more expensive network to salinient parts of
the input image for better prediction. Our work processes images as
a whole and concentrates on feature sharing instead.

\textbf{Model Compression.} The research on speeding up the evaluation
of DNNs is related in general, especially the works exploring redundancy
in networks. Knowledge distillation \cite{Distilling,FitNets} aims
to compress models in a student-teacher framework, whereas matrix
factorization methods \cite{Jaderberg2014,Denton2014,Zhang15} replace
weight matrices by their low-rank approximations. Computational efficiency
can be also incorporated from the beginning by imposing \eg a special
filter structure \cite{CircProj} or sparse filter connectivity \cite{DeepRoots}.
In a sense, we also exploit redundancy present in our baselines, assuming
it is possible to separate a certain amount of layers/channels into
an individual stage, which still performs reasonably well on the same
(sub)task. However, our motivation is different, we train our cascade
concurrently from scratch with the aim to use the full, combined network
as the last stage as well.

\vspace{-0.01\textheight}

\section{Method}

Our model is a cascade of feed-forward DNNs, called stages, evaluated
sequentially at test time. The aim of the cascade is to confidently
discriminate an input example as early in the classifier pipeline
as possible, saving running time. To deal with gradually more complicated
examples, the later stages should be more refined and operate on a
higher level of abstraction. 

\textbf{Motivation.} Multiple networks of different sizes trained
on the same dataset and for the same or a similar objective raise
the question whether their learned features have something in common.
Li \etal \cite{ConvLearning} confirm this for the case of different
initializations of the same network. Our major assumption is that
the feature maps at a particular layer computed by a smaller network
can be approximately subsumed by the feature maps of a larger network.
Thus, the larger network can be seen as an envelope around the smaller
network, adding new feature channels or layers and (partially) reusing
the features of the smaller network. Specifically, each convolutional
layer of the larger network receives the respective feature maps from
all smaller networks as an additional input. The key observation is
that these are shared and don't have to be recomputed. Our cascade,
coined OnionNet, can be pictured as an onion, each next stage wrapping
the previous. 

\textbf{Depth vs.~Width.} A natural way of constructing such a cascade
might be to gradually increase the depth only.  This is principally
similar to training a deeply supervised network \cite{Lee14} and
proceeding to deeper layers at test time until an associated 'local
companion output' rejects the example. However, the first layers are
the most expensive to compute due to large spatial size \cite{HeSun14}
while tending to produce weak classifiers due to few non-linearities
\cite{Vgg14}. Instead, we assume it is likely that early stages of
the cascade don't need as many feature maps as the later ones, which
leads us to construct the cascade by gradually increasing the width,
possibly in addition to depth. Making a stage thin reduces the burden
significantly and permits the cascade to delay fully evaluating expensive
lower layers until necessary.

In the rest of the paper, we restrict our scope to a two-stage cascade
only. However, our approach can be easily generalized to cascades
with more stages.

\vspace{-0.01\textheight}

\subsection{Model Description\label{sub:Model_OnionNet}}

Our two-stage OnionNet cascade consists of two branches with the same
layer organization (Figure~\ref{fig:method_overview}). Each takes
the same input and is terminated by its own output layer. The core
idea is that the branches are linked before every convolutional layer,
including the final one. The feature maps of the first stage (S1)
are used as additional input to the following convolutional layer
in the second stage (S2) but not the other way round, creating a one-way
dependence. Let $n_{l}^{N}$ denote the number of filters in the $l$-th
convolutional layer $C_{l}^{N}$ of network $N$. Then $C_{l}^{S2}$
receives its combined input of size $n_{l-1}^{S1}+n_{l-1}^{S2}$ from
both S2 and S1 and, conversely, the output of the layer immediately
preceeding $C_{l}^{S1}$ (usually a ReLU or a max-pooling layer) is
forwarded to both $C_{l}^{S1}$ and $C_{l}^{S2}$. 

In applications, OnionNet is designed as a replacement for a large
monolithic network $N_{\mathtt{M}}$. A simple way to configure a
cascade is to keep the effective number of filters per layer  unchanged,
i.e. splitting $n_{l}^{M}$ filters of $C_{l}^{M}$ to $n_{l}^{S1}$
and $n_{l}^{S2}$ filters, where $n_{l}^{S1}+n_{l}^{S2}=n_{l}^{M}$.
Although the number of feature maps is preserved, the amount of weights
decreases due to missing connections from S2 to S1 by $s_{l}^{2}n_{l-1}^{S2}n_{l}^{S1}$,
where $s_{l}$ denotes the size of filters (common to all $N_{\mathtt{M}}$,
S1, and S2). This has the same, albeit less severe effect on both
speed and accuracy as the so-called filter groups, which arise when
splitting layers among multiple GPUs \cite{Krizhevsky12} or by imposing
structure-induced regularization \cite{DeepRoots}.

\vspace{-0.01\textheight}

\subsection{Training}

Each stage is assigned its own loss function $L^{N}$, evaluated on
the output layer. Whereas $L^{S2}$ is application dependent, $L^{S1}$
is the standard cross-entropy loss over set of S1-classes $\mathcal{K}$.
OnionNet is trained jointly as a single model under the combined loss
$L=\alpha L^{S1}+(1-\alpha)L^{S2}$, where $\alpha\in(0,1)$ is a
fixed hyperparameter, each stage having access to the full training
set. Due to feature map sharing between the branches, the weights
of S1 (except for the last layer) receive backpropagation updates
from both $L^{S1}$ and $L^{S2}$, while the weights of S2 are trained
under $L^{S2}$ only. Therefore, the major benefit of joint training
is that the cascade learns the allocation of features between the
networks guided by the ratio of individual losses. In our initial
experiments with stage-wise independent training, we observed decreased
accuracy of S2 and an increased need for technical tweaks for it to
properly converge.

\vspace{-0.01\textheight}

\subsection{Testing}

\textbf{Thresholds.} We fix desired true positive rates (TPRs, recall)
for S1-classes of user interest $\mathcal{U\subset K}$, as we care
to precisely control the final accuracy rather than speed (false positive
rate, FPR). In order to choose such thresholds in a principled manner,
ROC curve is computed for each S1-class based on the score statistics
over the complete training set. A test example passes S1 if it scores
above any of $|\mathcal{U}|$ predefined thresholds, otherwise it
is rejected

\textbf{Sparse Batches.}  Not all examples in a test batch may pass
the first stage, leaving an irregular pattern of holes for S2. Unfortunately,
these cannot be easily skipped as no current GPU backend can work
with irregularly strided memory blocks. Thus, the batch as well as
all shared feature maps have to be reshuffled into smaller contiguous
blocks and the output of S2 scattered back. Formalized as obtaining
a single contiguous subsequence of 1s in a binary vector with the
least amount of move operations, this classic problem is solved efficiently
in just two passes over the vector.

\vspace{-0.01\textheight}

\section{Evaluation}

In this section, we evaluate OnionNet in three different applications:
descriptor matching, object detection, and image retrieval. 

In each application we compare the best-performing OnionNet cascade
$\mathit{\mathrm{\mathtt{\mathit{N}}}}$ to its respective monolithic
baseline network $N_{\mathtt{M}}$. The configurations, listed in
Table~\ref{tab:Architectures}, are designed to keep the effective
number of filters in S2 as in $N_{\mathtt{M}}$ (Section~\ref{sub:Model_OnionNet}).
The non-sharing cascade $\mathit{\mathrm{\mathtt{\mathit{N}_{NS}}}}$
constitutes our second baseline; its stages do not share features
but have the same number of effective filters in each stage as $\mathit{\mathrm{\mathtt{\mathit{N}}}}$.

Implementation was done in Torch \cite{Torch} with cuDNN backend
\cite{cudnn} with auto-tuning to use the fastest convolution algorithms.
Mean running time over 50 executions on NVIDIA Titan Black is reported
with its standard error. We time solely the forward pass, and not
\eg any preprocessing or uploading of the batch. In addition, we
report $\bar{p}$ as the mean percentage $p$ of examples in a batch
passing S1, which is indicative of the strength of the first stage. 

\begin{table*}
\begin{centering}
\resizebox{0.8\linewidth}{!}{%
\begin{tabular}{|l|l|}
\hline 
$\mathtt{P_{M}}$ \cite{SZ15} &
C4(96), 3\texttimes C3(96), 3\texttimes C3(192), C2(1)\tabularnewline
$\mathrm{\mathtt{P}}$ &
C4(32/64), 3\texttimes C3(32/64), 3\texttimes C3(64/128), C2(2/1)\tabularnewline
\hline 
$\mathtt{D_{M}}$ \cite{Krizhevsky12} &
C11(96), C5(256), 2\texttimes C3(384), C3(256), C6(4096), C1(4096),
C1(21)\tabularnewline
$\mathrm{\mathtt{D}}$ &
C11(96), C5(256), 2\texttimes C3(384), C3(256), C6(512/3584), C1(512/3584),
C1(2/21)\tabularnewline
\hline 
$\mathtt{R_{M}}$ \cite{Chatfield14} &
C11(64), C5(256), 3\texttimes C3(256), C6(4096), C1(4096), C1(4096)\tabularnewline
$\mathrm{\mathtt{R}}$ &
C11(8/56), C5(32/224), 3\texttimes C3(32/224), C6(256/3840), C1(256/3840),
C1(7/1000)\tabularnewline
\hline 
\end{tabular}}
\par\end{centering}

\centering{}\medskip{}
\caption{\label{tab:Architectures}Configuration of the monolitic baselines
($\mathtt{P_{M}}$ for patch comparison, $\mathtt{D_{M}}$ for object
detection, and $\mathtt{R_{M}}$ for image retrieval) and their best
OnionNet models ($\mathtt{P}$, $\mathtt{D}$, $\mathtt{R}$). Only
parametric layers are listed for clarity, other layers and parameters
are consistent with the paper having introduced the baseline. Fully
connected layers are implemented as convolutions. C$s$($n$) denotes
a convolutional layer with $n$ output filters of spatial size $s\times s$.
In OnionNets, C$s$($n^{S1}/n^{S2}$) denotes a pair of convolutional
layers: C$s$($n^{S1}$) in S1 and C$s$($n^{S2}$) in S2. }
\vspace{-0.01\textheight}
\end{table*}

\vspace{-0.01\textheight}

\subsection{Application: Comparing Patches\label{sec:AppPatch}}

DNNs have been applied to comparing patches just recently, achieving
state-of-the-art results. While the ultimate goal might be to learn
L2 embeddings of deep descriptors \cite{Serra15}, comparing descriptor
pairs using a matching network \cite{MatchNet15,SZ15,Zbontar15} and
particularly processing patch pairs jointly from the start were shown
to be the best-performing solutions so far \cite{SZ15}. However,
as there are quadratically many pairs and the joint (sub)network has
to be evaluated for each comparison, such architectures seem rather
impractical. Fortunately, the expected high number of easy negative
pairs makes for a natural application of OnionNet. This holds especially
true for feature point matching between images.  

\textbf{Setting.} We evaluate on two datasets. The multi-view stereo
correspondence dataset (MVSD) of Brown \etal \cite{BrownPatches11}
is a balanced dataset of grayscale patches. There are three subsets;
we train on Notre Dame and test on Liberty and Yosemite, reporting
the false positive rate at 95\% recall as in \cite{SZ15}. The local
descriptor benchmark (LDB) of Mikolajczyk and Schmid \cite{Mikolajczyk}
consists of 6 images sequences with ground truth homographies. This
dataset is expected to have an unbalanced, realistic proportion of
positives and negatives. We use the framework of \cite{lenc12vlbenchmarks}
for evaluation, regions of interest being extracted using MSER detector. 

To fully demonstrate our advantage, we experiment with ``2ch-deep''
model $\mathtt{P_{M}}$ from Zagoruyko and Komodakis \cite{SZ15}.
Note that our method could also be applied to matching networks in
the same way. All models were trained from scratch for 256 epochs
with $L^{S2}$ being binary hinge loss and $\alpha=0.5$. ASGD with
learning rate 0.1, weight decay 0.0005, and momentum 0.9 was used
to train the models with batch size 128 and random flipping.

\begin{figure*}

\begin{minipage}[t]{0.52\textwidth}%
\begin{table}[H]
\resizebox{1\linewidth}{!}{%
\begin{tabular}{|l|c|c|c|}
\cline{2-4} 
\multicolumn{1}{l|}{} &
\multicolumn{1}{c|}{FPR@95} &
\multicolumn{1}{c|}{$\bar{p}$} &
\multicolumn{1}{c|}{running time {[}sec{]}}\tabularnewline
\hline 
$\mathtt{P_{M}}$  &
0.0503 &
100.00\%  &
12.111 \textpm{} 0.001 \tabularnewline
$\mathrm{\mathtt{P_{NS}}}$  &
0.0514 &
55.44\% &
9.677 \textpm{} 0.005 \tabularnewline
$\mathrm{\mathtt{P}}$  &
0.0601 &
54.05\% &
8.407 \textpm{} 0.006 \tabularnewline
\hline 
\end{tabular}}

\begin{centering}
\medskip{}

\par\end{centering}

\caption{\label{tab:LibYos}Descriptor matching (MVSD). Average over Liberty
and Yosemite subsets at TPR=0.99 on S1. FPR@95 is FPR at TPR=0.95
on S2/baseline, $\bar{p}$ is mean percentage of examples passing
S1, running time is normalized per 100K examples.}
\end{table}
\end{minipage}\hfill{}%
\begin{minipage}[t]{0.48\textwidth}%
\begin{figure}[H]
\begin{centering}
\vspace{-0.01\textheight}
\resizebox{0.8\linewidth}{!}{\includegraphics[width=0.65\linewidth]{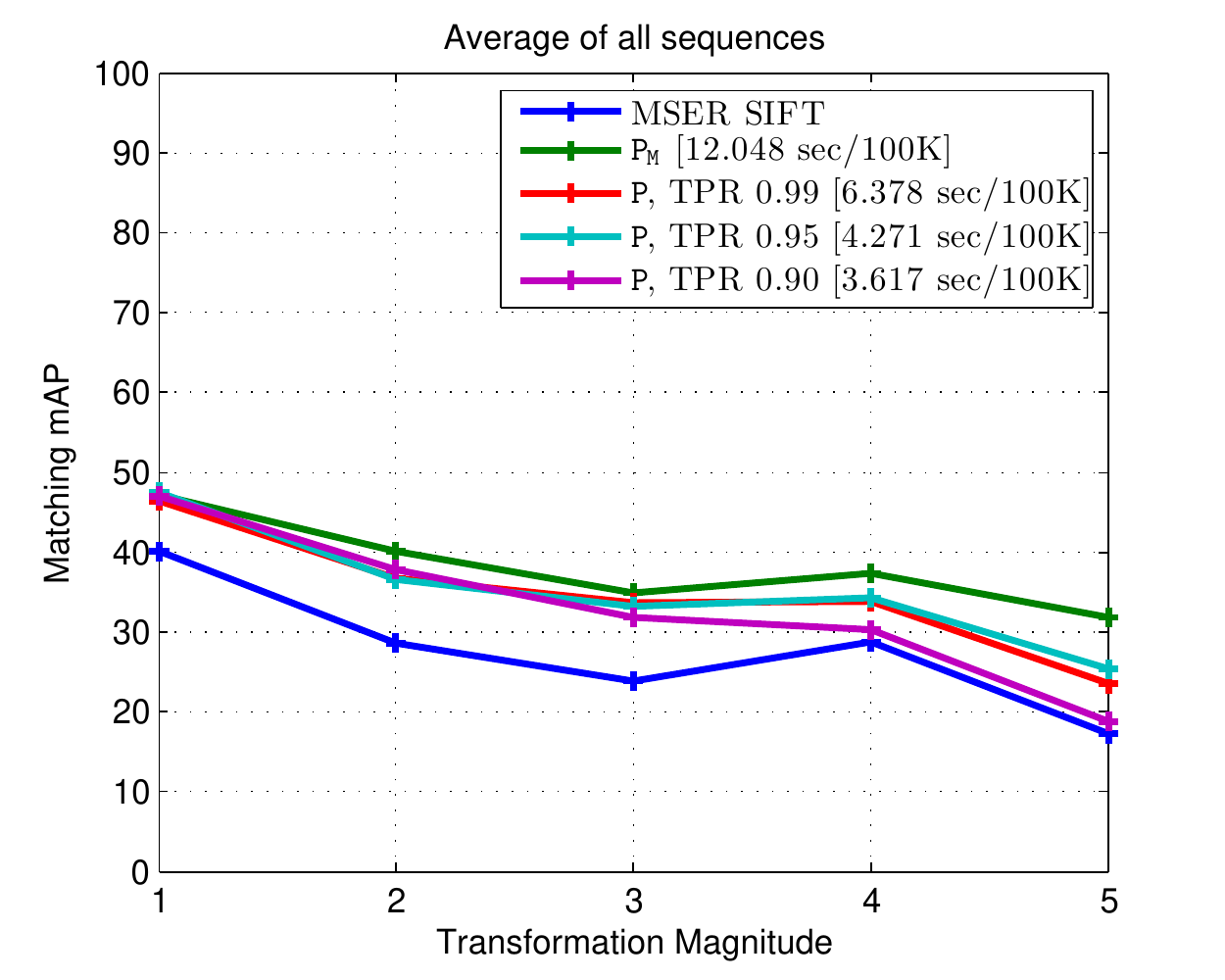}}\vspace{-0.013\textheight}

\par\end{centering}

\caption{\label{fig:Mikol}Descriptor matching (LDB).}
\end{figure}
\end{minipage}

\vspace{-0.013\textheight}

\end{figure*}

\textbf{Results.} Table~\ref{tab:LibYos} lists the results on MVSD
with TPR of S1 set to 0.99\footnote{Setting S1-TPR to 0.98 already made it produce enough false negatives
so that the prescribed S2-TPR of 0.95 was never reached. This is mostly
due to domain transfer, as the particular threshold for S1 was chosen
based on ROC curve computed on the training set of a different subset.}. OnionNet outperforms its baselines in terms of running time (by
31\% for $\mathtt{P_{M}}$ and 13\% for $\mathrm{\mathtt{P_{NS}}}$).
The better mAP but worse running time of $\mathrm{\mathtt{P_{NS}}}$
\wrt $\mathtt{P}$ is justified in Section~\ref{sub:Tradeoff-analysis}.
Figure~\ref{fig:Mikol} plots mAP and running times on LDB averaged
over all types of transformations in the dataset. The evaluation of
LDB does not constrain us from choosing a larger set of TPRs: 0.99,
0.95, and 0.90. The plot shows that under a more realistic imbalance
of positive and negative examples we can archive considerable speedup
(up to 2.8x) with limited decrease of precision, which starts to show
up mostly under higher transformation magnitude. This is likely caused
by discarding positives difficult due their extreme deformation, which
places them near the decision boundary.

\vspace{-0.01\textheight}

\subsection{Application: Proposal-based Object Detection}

The currently dominant paradigm in object detection is to use an algorithm
to generate a set of object proposals, which are then verified by
a classifier. Object proposal algorithms are typically tuned for high
recall and are often class-agnostic, which allows them to be used
as an off-the-shelf preprocessing step. This flexibility comes at
a price of the classifier having to process many proposals that are
of no interest with respect to the task-specific set of classes. For
example, Fast R-CNN sifts through thousands of proposals per image~\cite{frcnn}.
Motivated by this, we propose to construct the classifer as OnionNet
so that its first stage serves as a background classifier, leaving
the task of identifying the particular classes to the second stage.
Note that such a task-specific scoring can be alternatively built
into the proposal generator itself, as demonstrated by several very
recent works~\cite{faster,DeepMask}.

\textbf{Setting.} We experiment with Fast R-CNN~\cite{frcnn} on
PASCAL VOC 2007 with precomputed Selective Search proposals available
at the author's webpage. The baseline $\mathrm{\mathtt{D_{M}}}$ is
their 'small' AlexNet network (our reproduction scores 0.023 mAP less).
Our proposal $\mathrm{\mathtt{D}}$ is created by replacing the classifier,
composed of 3 fully-connected layers, by OnionNet of the same width,
see Table~\ref{tab:Architectures}.  The classifier layers were
initialized randomly in all models and the whole networks were finetuned
on trainval subset of VOC 2007 for 120k iterations using SGD with
$\alpha=0.5$, $L^{S2}$ as cross-entropy loss, and learning rate
0.001, dropping to 0.0001 after 100k iterations. Bounding box regression
was omitted in the implementation \cite{fmassaOD}, which does not
affect conclusions from our comparison, though. We report mean average
precision (mAP) with S1 TPR fixed at 0.95.

\textbf{Results.} The results are listed in Table~\ref{tab:ObjDetRes-1}.
OnionNet $\mathrm{\mathtt{D}}$ is able to achieve 2.9x gain in speed
\wrt baseline $\mathtt{D_{M}}$ under a graceful degradation of 0.018
mAP points, which is better than the 1.72\texttimes{} speed-up attained
by SVD of the weight matrices as in \cite{frcnn}. Note that both
methods might be combined as they are basically orthogonal. We also
marginally outperform $\mathrm{\mathtt{D_{NS}}}$ in both running
time and precision.

\vspace{-0.01\textheight}

\subsection{Application: Image Retrieval\label{sec:AppRetr}}

We are motivated by retrieval over ephemeral datasets, where an index
building stage typical for image retrieval \cite{Jegou12} may be
too heavy; consider \eg a robot actively searching for a particular
object or a user wanting to copy images of only cats from his camera.
Instead, on-the-fly retrieval \cite{ChatfieldOTF14} casts such a
problem as classification. However, similar to the way human search,
the system does not need to precisely label every object it knows
unless it is the object being searched for. We demonstrate that designing
the classifier as OnionNet can lead to a significant decrease in running
time.

\textbf{Setting.} We train and validate on ILSVRC 2012 \cite{Imagenet}.
Although not perfectly suited for retrieval scenarios due to incomplete
annotations \cite{ChatfieldOTF14}, we choose it because of its scale
and our concentration on quantifying relative performance improvements.
We aim for retrieving images of a certain class from the set of 50k
validation images, rather than classifying all images. Therefore,
as in PASCAL VOC classification task, we report mean average precision
(mAP) over 1000 classes instead of accuracy. A test example is considered
retrieved if its true class is predicted within the top-5 softmaxed
scores. TPR of S1 is fixed at 0.9. S2-classes are partitioned into
7 S1-classes $\mathcal{K}$ by k-means clustering of class-averaged
activation.

Experiments are performed with Alexnet-like 'CNN-F' baseline $\mathtt{R_{M}}$
from \cite{Chatfield14}. All models were trained from scratch for
53 epochs with $L^{S2}$ being 1000-way cross-entropy loss and $\alpha=0.5$.
SGD with learning rate 0.01 (reduced to 0.005, 0.001, 0.0005, 0.0001
after 18, 29, 43, 52 epochs), weight decay 0.0005 for 29 epochs, and
momentum 0.9 was used to train the models with batch size 128. As
a reference, our $\mathtt{R_{M}}$ achieves 19.3\% top-5 error with
10 crops on the standard ILSVRC classification task. The S1 network
of our proposal $\mathrm{\mathtt{R}}$ is as deep as $\mathtt{R_{M}}$
and contains $\nicefrac{1}{8}$, resp. $\nicefrac{1}{16}$ of its
convolutional, resp. fully-connected filters, see Table~\ref{tab:Architectures}.

\textbf{Results.} The results are listed in Table~\ref{tab:ImRetRes-1}.
OnionNet is able to cut the running time by 41.5\% while giving up
only 0.049 mAP points \wrt baseline $\mathtt{R_{M}}$. It also saves
7\% time \wrt non-sharing cascade $\mathrm{\mathtt{R_{NS}}}$, which
is a fair result given the relatively low amount of shared feature
maps. The better mAP but worse running time of $\mathrm{\mathtt{R_{NS}}}$
\wrt $\mathtt{R}$ is justified in Section~\ref{sub:Tradeoff-analysis}.

\begin{figure*}

\begin{minipage}[t]{0.48\textwidth}%
\begin{center}
\begin{table}[H]
\begin{centering}
\resizebox{1\linewidth}{!}{%
\begin{tabular}{|l|c|c|c|}
\cline{2-4} 
\multicolumn{1}{l|}{} &
\multicolumn{1}{c|}{mAP} &
\multicolumn{1}{c|}{$\bar{p}$} &
\multicolumn{1}{c|}{running time {[}sec{]}}\tabularnewline
\hline 
$\mathtt{R_{M}}$  &
0.587 &
100\% &
49.271 \textpm{} 0.117 \tabularnewline
$\mathrm{\mathtt{R_{NS}}}$ &
0.551 &
33.52\% &
30.991 \textpm{} 0.307\tabularnewline
$\mathrm{\mathtt{R}}$ &
0.538  &
32.56\% &
28.806 \textpm{} 0.268\tabularnewline
\hline 
\end{tabular}}
\par\end{centering}

\begin{centering}
\medskip{}

\par\end{centering}

\caption{\label{tab:ImRetRes-1}Image retrieval (ILSVRC 2012) at TPR=0.9 on
S1, $\bar{p}$ is mean \% of examples passing S1, time is per dataset.}
\end{table}

\par\end{center}%
\end{minipage}\hfill{}%
\begin{minipage}[t]{0.47\textwidth}%
\begin{center}
\begin{table}[H]
\begin{centering}
\resizebox{1\linewidth}{!}{%
\begin{tabular}{|l|c|c|c|}
\cline{2-4} 
\multicolumn{1}{l|}{} &
\multicolumn{1}{c|}{mAP} &
\multicolumn{1}{c|}{$\bar{p}$} &
\multicolumn{1}{c|}{running time {[}ms{]}}\tabularnewline
\hline 
$\mathrm{\mathtt{D_{M}}}$  &
0.499 &
100\% &
96.711 \textpm{} 0.010\tabularnewline
$\mathrm{\mathtt{D_{NS}}}$ &
0.479 &
2.71\% &
33.752 \textpm{} 0.008\tabularnewline
$\mathrm{\mathtt{D}}$ &
0.481 &
2.65\% &
33.258 \textpm{} 0.003\tabularnewline
\hline 
\end{tabular}}
\par\end{centering}

\begin{centering}
\medskip{}

\par\end{centering}

\caption{\label{tab:ObjDetRes-1}Object detection (VOC 2007) at TPR=0.95 on
S1, $\bar{p}$ is mean \% of examples passing S1, time is per example.}
\end{table}

\par\end{center}%
\end{minipage}

\vspace{-0.055\textheight}

\end{figure*}\vspace{-0.01\textheight}

\section{Discussion}

In this section we conduct further analysis in order to gain insight
into the properties of OnionNet. To that end, we define multiple variants
of OnionNet for the image retrieval network $\mathtt{R_{M}}$ by varying
the width or depth of S1. We extend our notation by superscripts for
that: the S1 network of $\mathrm{\mathtt{\mathit{\mathtt{R}}^{W\mathit{w}}}}$
has the width of grade $w$ (the greater the wider), the depth $d$
of $\mathrm{\mathtt{\mathit{\mathtt{R}}^{D\mathit{d}}}}$ being denoted
accordingly. First, we study how the ratio of number of filters allocated
to S1 and S2 influences the overall performance. Second, we define
a theoretical time complexity and compare it to the empirical running
time on a synthetic benchmark.

\vspace{-0.01\textheight}

\subsection{Trade-off Analysis\label{sub:Tradeoff-analysis}}

We analyze the effect of reducing the depth or width of $\mathrm{\mathtt{R_{M}}}$
by evaluating networks listed in Table~\ref{tab:WidthDepth}. The
results convey that this makes S1 weaker, as measured by $\bar{p}$,
and S2 stronger, as measured by mAP when S1 is deactivated and all
examples pass it (column 'full'). This is expected, as the accuracy
of a network is highly dependent on the amount of allocated filters
and parameters. Regarding running time, depth reduction brings less
benefit than that of width due to lower layers being the most expensive
to compute: $\mathrm{\mathtt{R^{D2}}}$, mimicking a Faster R-CNN-like
cascade, can spare only 6.8\% time. 

In general, neither sharing nor non-sharing cascades are expected
to reach the accuracy of their monolithic baseline due to a non-zero
false negative rate at S1. Sharing cascades trade even more accuracy
for speed by parameter reduction in S2 (Section~\ref{sub:Model_OnionNet})
and shared features serving two different objectives. It can therefore
be observed in the column 'full' that none of the OnionNet cascades
can achieve the mAP of $\mathrm{\mathtt{R_{M}}}$. To confirm the
effect of joint learning, we increased the importance of S2 by retraining
$\mathrm{\mathtt{R}}$ with $\alpha=0.25$ and obtained improvement
of around 0.015 mAP points at an increase of $\bar{p}$ of around
1.5\% points.

\begin{table*}
\begin{centering}
\resizebox{\linewidth}{!}{%
\begin{tabular}{|ll|cc|c|c|}
\cline{3-6} 
\multicolumn{2}{c|}{} &
\multicolumn{2}{c|}{mAP} &
\multicolumn{1}{c|}{$\bar{p}$} &
\multicolumn{1}{c|}{running time {[}sec{]}}\tabularnewline
\multicolumn{2}{r|}{TPR} &
full &
0.9 &
0.9 &
0.9\tabularnewline
\hline 
$\mathtt{R_{M}}$ &
C11(64), C5(256), 3\texttimes C3(256), C6(4096), C1(4096), C1(4096) &
0.587  &
0.587  &
100.00\% &
49.271 \textpm{} 0.117\tabularnewline
$\mathrm{\mathtt{R^{W3}}}$ &
C11(32/32), C5(128/128), 3\texttimes C3(128/128), C6(1024/3072), C1(1024/3072),
C1(7/1000) &
0.550  &
0.524  &
21.38\%  &
29.864 \textpm{} 0.191\tabularnewline
$\mathrm{\mathtt{R^{W2}}}$ &
C11(16/48), C5(64/192), 3\texttimes C3(64/192), C6(512/3584), C1(512/3584),
C1(7/1000) &
0.558  &
0.528  &
26.34\%  &
28.431 \textpm{} 0.241\tabularnewline
$\mathrm{\mathtt{R^{W1}}=\mathtt{R}}$ &
C11(8/56), C5(32/224), 3\texttimes C3(32/224), C6(256/3840), C1(256/3840),
C1(7/1000) &
0.573  &
0.538  &
32.56\%  &
28.806 \textpm{} 0.268\tabularnewline
$\mathrm{\mathtt{R^{D2}}}$ &
C11(64/-), C5(256/-), 3\texttimes C3(256/-), C6(7/4096), C1(-/4096),
C1(-/1000) &
0.565  &
0.536  &
20.76\%\textbf{ } &
45.923 \textpm{} 0.189\tabularnewline
$\mathrm{\mathtt{R^{D1}}}$ &
C11(64/-), C5(256/-), C3(256/-), C3(7/256), C3(-/256), C6(-/4096),
C1(-/4096), C1(-/1000) &
0.493  &
0.470  &
24.35\%  &
40.448 \textpm{} 0.220\tabularnewline
\hline 
\end{tabular}}
\par\end{centering}

\begin{centering}
\medskip{}

\par\end{centering}

\caption{\label{tab:WidthDepth}Image retrieval (ILSVRC 2012) with S1 networks
of various width and depth. TPR 'full' allows every example to pass
S1. The configuration notation is as in Table~\ref{tab:Architectures}.
Further, if S2 starts deeper or S1 ends shallower, missing layers
are indicated by ``-'' and S1 networks then contain an extra max-pooling
layer before their final convolutional layer.}
\end{table*}

\vspace{-0.01\textheight}

\subsection{Time Cost Analysis\label{sec:Time-cost-analysis}}

\begin{figure*}
\begin{centering}
\includegraphics[width=0.33\linewidth]{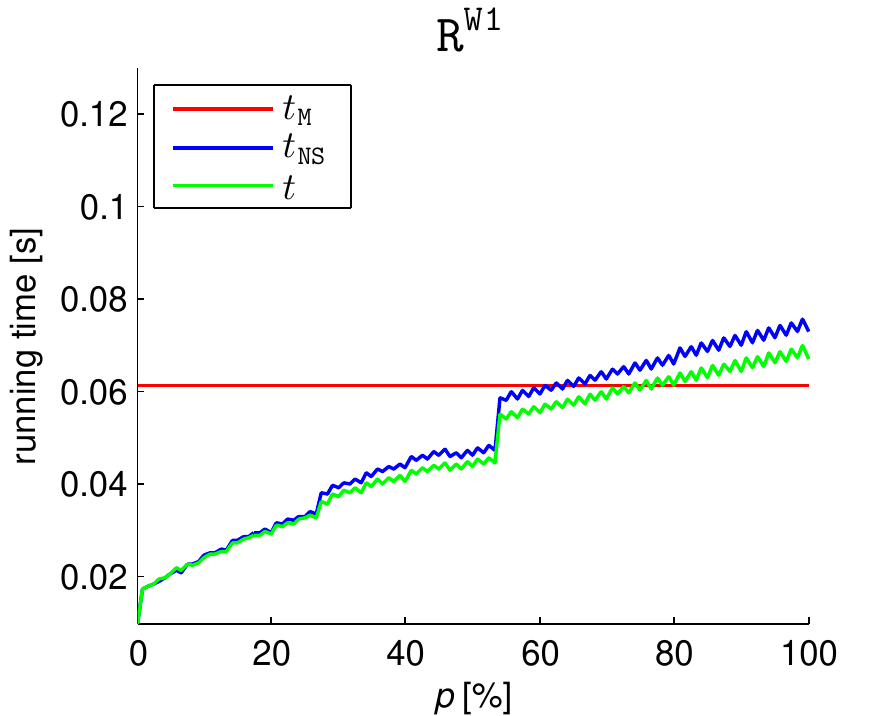}\includegraphics[width=0.33\linewidth]{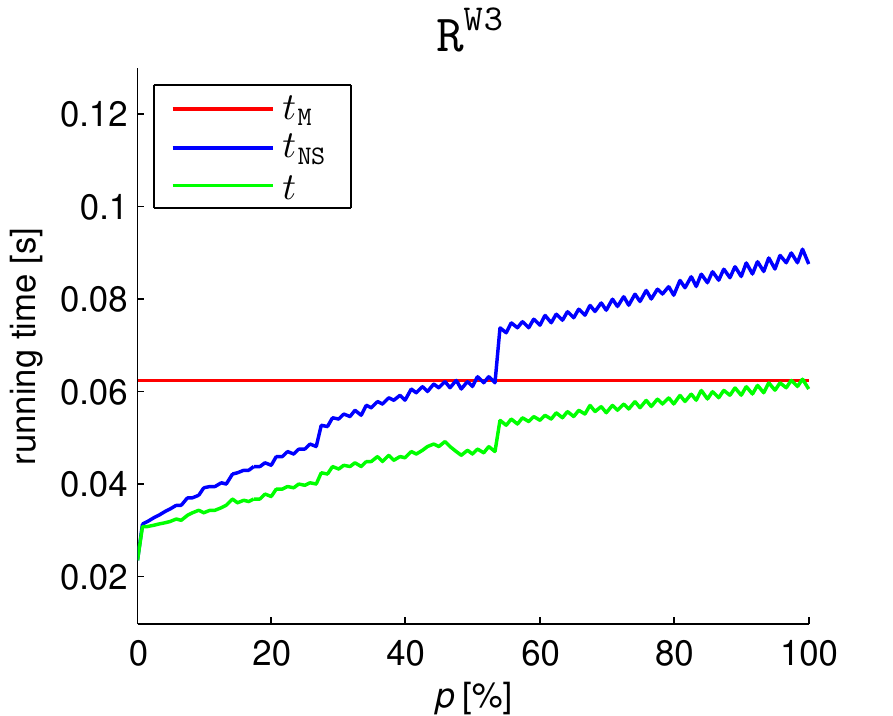}\includegraphics[width=0.33\linewidth]{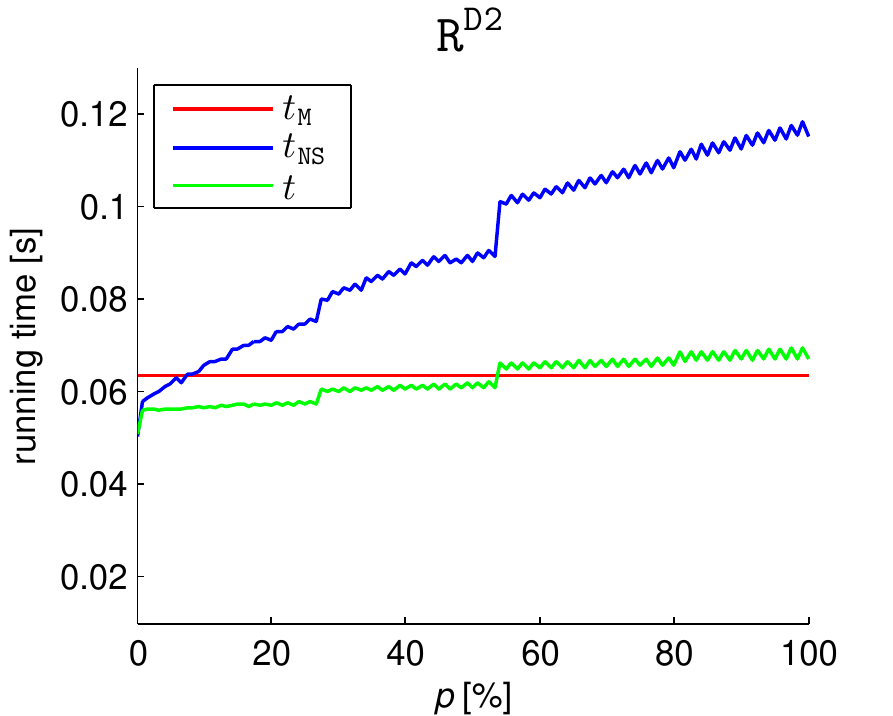}
\par\end{centering}

\begin{centering}
\includegraphics[width=0.33\linewidth]{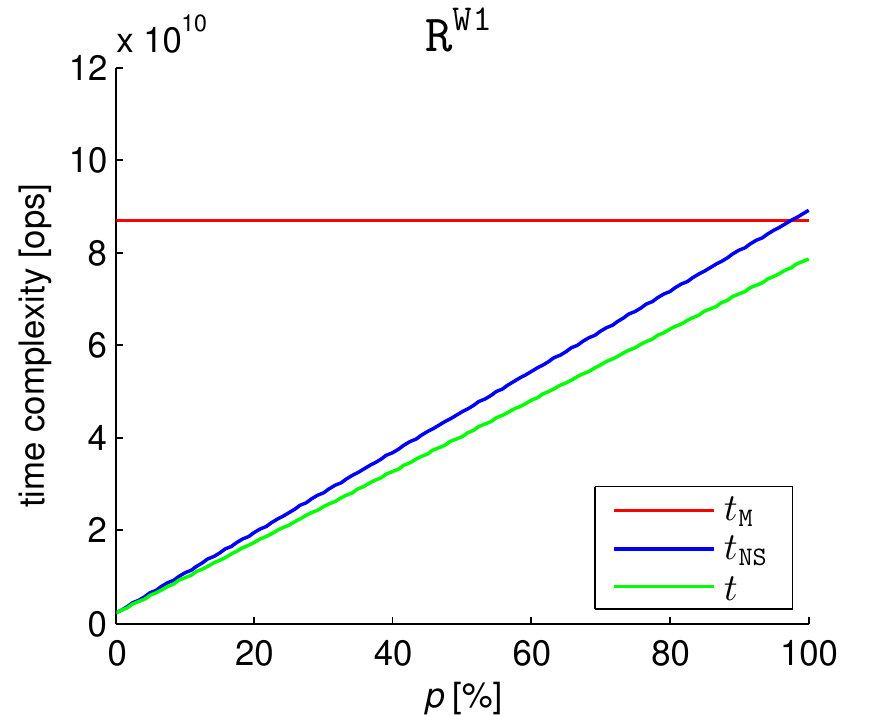}\includegraphics[width=0.33\linewidth]{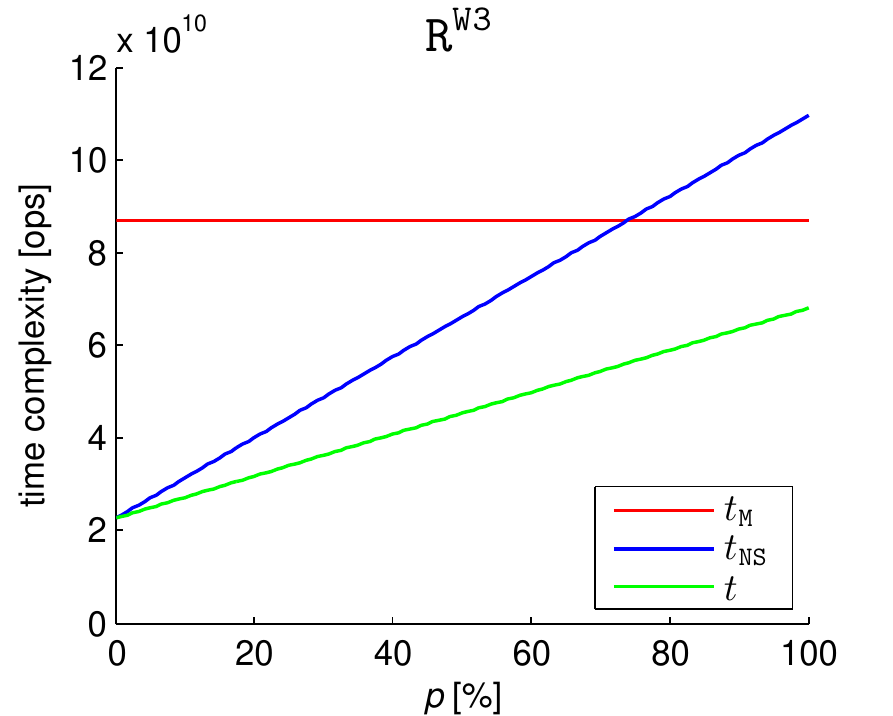}\includegraphics[width=0.33\linewidth]{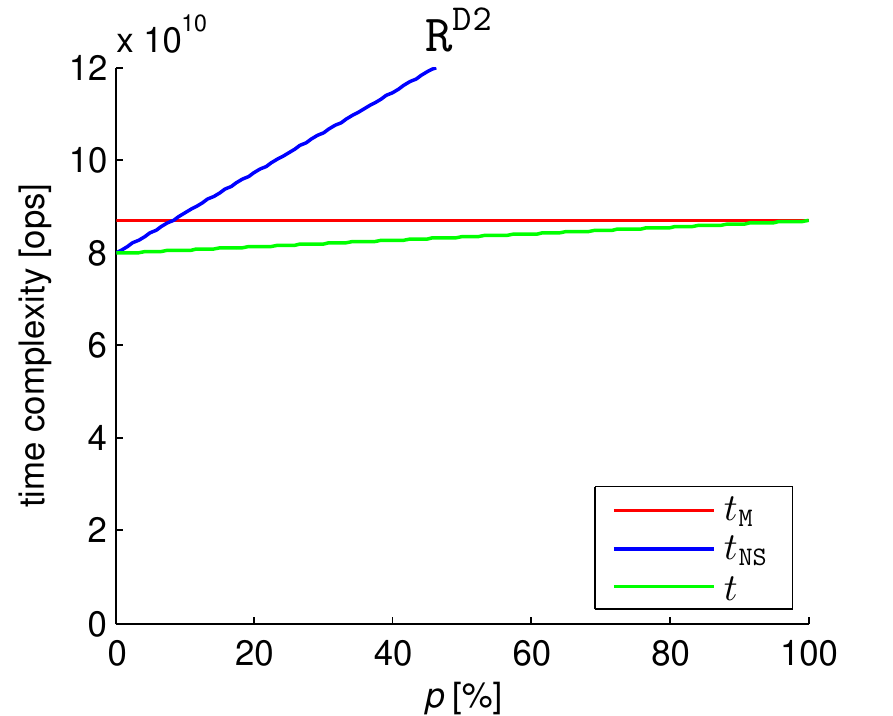}
\par\end{centering}

\caption{\label{fig:TimeCosts}Empirical running times (first row) and time
complexities (second row) for $\mathrm{\mathtt{R_{M}}}$-based cascades
as functions of the percentage $p$ of examples passing S1. Sharing
cascade ($t$) is compared with a corresponding non-sharing baseline
($t_{\mathtt{NS}}$) and its monolithic baseline ($t_{\mathtt{M}})$.}
\vspace{-0.013\textheight}
\end{figure*}

While measuring empirical running time makes for a practical comparison,
its generality is limited due to inherent sensitivity to system (esp.
GPU architecture) and implementation factors (esp. DNN backend). Thus,
we additionally investigate the theoretical time complexity as introduced
by He and Sun~\cite{HeSun14}. The total time complexity of convolutional
layers is defined\footnote{This definition does not involve non-convolutional layers and batch
reshuffling before evaluating S2: pooling layers ``often take 5-10\%
computational time''\cite{HeSun14} and the theoretical complexity
of reshuffling is negligible.} in the notation of Section~\ref{sub:Model_OnionNet} as $O(\sum_{i=1}^{2}\sum_{l=1}^{d_{i}}n_{l-1}^{Si}s_{l}^{2}n_{l}^{Si}m_{l}^{2})$,
where $d_{i}$ is number of convolutional layers in a stage and $m_{l}$
is the spatial size of an output feature map.

The time cost behavior of a cascade can be best described as a function
of the percentage $p$ of examples in a batch passing S1. We plot
the costs $t(p)$, $t_{\mathtt{M}}(p)$, and $t_{\mathtt{NS}}(p)$
of networks $\mathrm{\mathtt{\mathit{N}}}$, $\mathrm{\mathtt{\mathit{N}}}_{\mathtt{M}}$,
and $\mathrm{\mathtt{\mathit{N_{\mathtt{NS}}}}}$ respectively on
a synthetic benchmark where we can regulate $p$ as necessary. The
disparity $t_{\mathtt{NS}}-t_{\mathtt{M}}$ shows for which $p$ a
cascade is actually useful and the disparity $t-t_{\mathtt{NS}}$
reveals the margin of OnionNet due to parameter reduction and feature
map sharing. Results for prominent networks of Table~\ref{tab:WidthDepth}
are shown in Figure~\ref{fig:TimeCosts} (batch size 120).

\textbf{Time~Complexity.} The plots suggests that OnionNet always
improves on time cost, as $\forall p:t<t_{\mathtt{NS}}$. Note that
$t_{\mathtt{NS}}>t_{\mathtt{M}}$ for higher values of $p$, \ie
non-sharing cascades are overperformed by the monolithic classifiers
at some point. The plots also convey that large S1 networks benefit
from the speed-up the most ($\mathrm{\mathtt{R^{W1}}}$ vs. $\mathrm{\mathtt{R^{W3}}}$).
Also, we can notice that S1 networks of unreduced width are very costly
($\mathrm{\mathtt{R^{D2}}}$) even for small $p$ values, despite
the heavy help from feature map sharing.

\textbf{Running~Time.} The plots follow the general trend of time
complexity plots, although with some important differences. We observe
that many configurations perform worse than their monolithic baseline
($t>t_{\mathtt{M}}$) for large $p$, except for the configuration
with the largest S1 networks ($\mathrm{\mathtt{R^{W3}}}$). Nevertheless,
the behavior under smaller $p$ values, \ie those reported in our
applications, appears still very promising. We have identified two
causes for such an inconsistency between theory and practice, also
reported by \cite{Denton2014,DeepRoots}. One is a nonlinear, nonmonotonous
relation of data size and convolution running time due to some sizes
being more 'GPU friendly'. The other is the overhead of layer executions,
esp. CuDNN kernel launches, since cascades have to basically perform
the forward pass twice.

To summarize the analysis, we have shown that our model is theoretically
well founded, although overhead of current GPU solutions has to be
considered in practice, which may render small weak classifiers ill-suited
for a cascaded solution in general. Due to similar reasons, using
more than two stages turned out impractical in our initial experiments.
Since the actual benefit of a cascade varies by $p$, which in practice
depends on the precision of S1 at a chosen true positive rate, it
was important to identify the sweet spots in practical applications,
as we successfully demonstrated above.

\vspace{-0.01\textheight}

\section{Conclusion}

\begin{figure}
\begin{centering}
\includegraphics[width=0.33\linewidth]{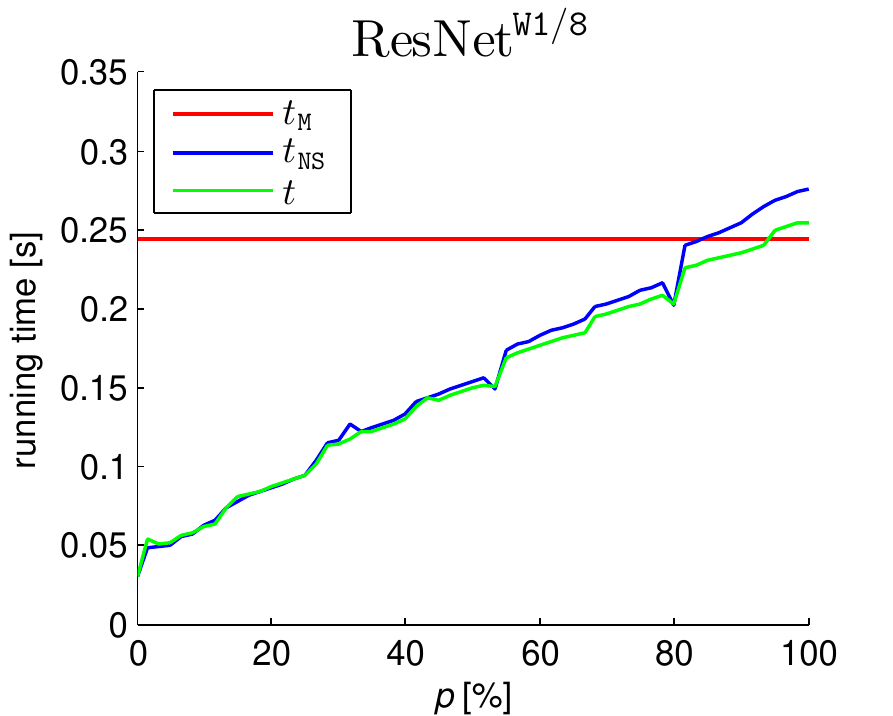}\includegraphics[width=0.33\linewidth]{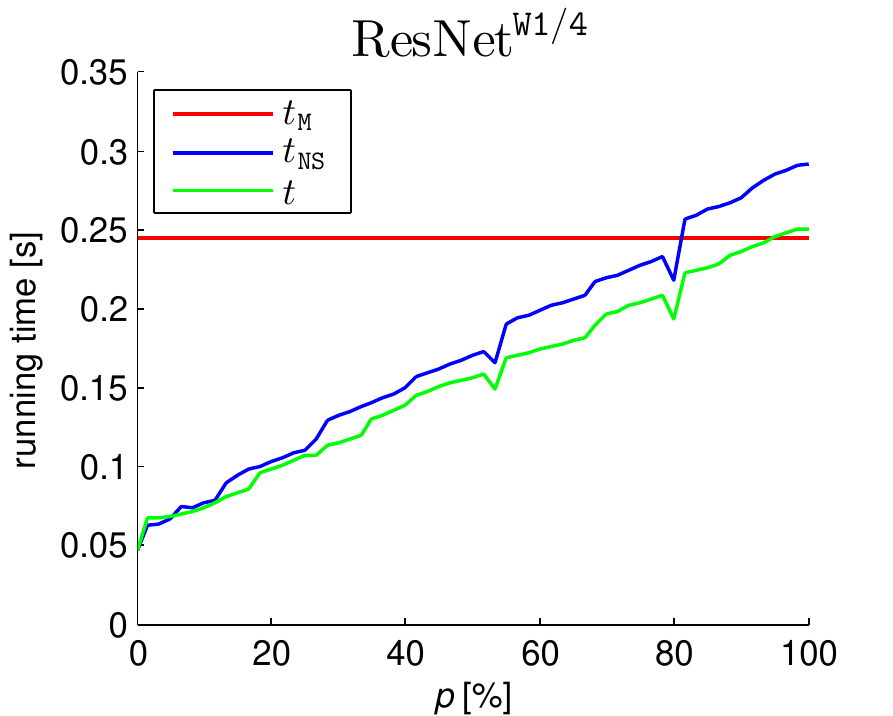}\includegraphics[width=0.33\linewidth]{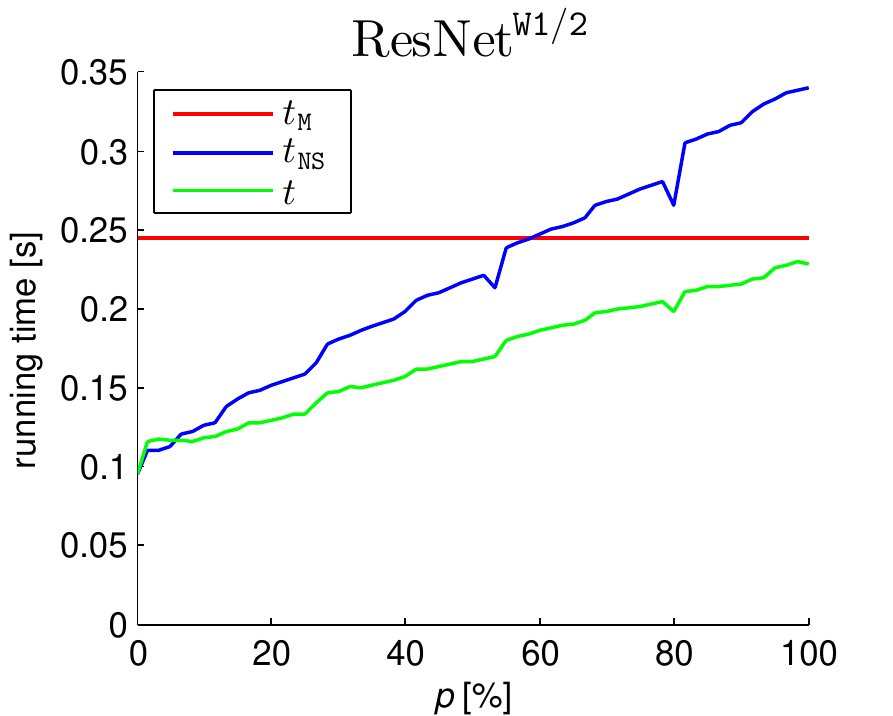}
\par\end{centering}

\caption{\label{fig:resnet}Empirical running times for cascades based on ImageNet
ResNet-34 B network \cite{residuals} created by allocating $\nicefrac{1}{8}$,
$\nicefrac{1}{4}$, or $\nicefrac{1}{2}$ of feature channels in each
convolutional layer to S1 and leaving the rest in S2; both stages
are of the same depth. The notation is identical to Figure~\ref{fig:TimeCosts}
and the conclusions from Section~\ref{sec:Time-cost-analysis} hold
here as well.}
\vspace{-0.013\textheight}
\end{figure}

A novel cascade of feature-sharing deep classifiers was proposed where
subsequent stages may be extended by new layers and/or feature channels
and their intermediate computations reused. Our motivation was to
speed up the evaluation by preventing similar features from being
recomputed, which led us to make each stage of the cascade equally
deep. Sharing and reduction in model parameters are the main causes
of the achieved speed-up. The same factors account for a minor decrease
in precision, though. We have demonstrated good speed-ups due to cascades
in three important tasks and showed that OnionNet sharing can bring
further gain atop of it. We find this fact encouraging, as our applications
seem to require some higher-level understanding even for the easy
examples, and thus massive speed-ups due to very simple stages as
in sliding-window methods should not be expected. As much deeper,
more expensive networks are being introduced \cite{residuals}, we
believe our method might gain in significance due to larger absolute
running time savings; see Figure~\ref{fig:resnet} for a preliminary
time cost analysis of a 34-layer residual network cascade.

\vspace{0.5ex}

\textbf{Acknowledgments. }We gratefully acknowledge NVIDIA Corporation
for the donated GPU used in this research and Sergey Zagoruyko for
providing his early source code for patch matching \cite{SZ15}. 

\bibliographystyle{plain}
\bibliography{bmvc_final}

\begin{thebibliography}{41}
\providecommand{\natexlab}[1]{#1}
\providecommand{\url}[1]{\texttt{#1}}
\expandafter\ifx\csname urlstyle\endcsname\relax
  \providecommand{\doi}[1]{doi: #1}\else
  \providecommand{\doi}{doi: \begingroup \urlstyle{rm}\Url}\fi

\bibitem[Almahairi et~al.(2015)Almahairi, Ballas, Cooijmans, Zheng, Larochelle,
  and Courville]{DynaCap}
Amjad Almahairi, Nicolas Ballas, Tim Cooijmans, Yin Zheng, Hugo Larochelle, and
  Aaron~C. Courville.
\newblock Dynamic capacity networks.
\newblock \emph{CoRR}, abs/1511.07838, 2015.

\bibitem[Angelova et~al.(2015)Angelova, Krizhevsky, Vanhoucke, Ogale, and
  Ferguson]{Angelova15}
Anelia Angelova, Alex Krizhevsky, Vincent Vanhoucke, Abhijit Ogale, and Dave
  Ferguson.
\newblock Real-time pedestrian detection with deep network cascades.
\newblock In \emph{BMVC}, 2015.

\bibitem[Bourdev and Brandt(2005)]{SoftCascade}
Lubomir Bourdev and Jonathan Brandt.
\newblock Robust object detection via soft cascade.
\newblock In \emph{CVPR}, 2005.

\bibitem[Brown et~al.(2011)Brown, Hua, and Winder]{BrownPatches11}
Matthew Brown, Gang Hua, and Simon A.~J. Winder.
\newblock Discriminative learning of local image descriptors.
\newblock \emph{IEEE Transactions on Pattern Analysis \& Machine Intelligence},
  33\penalty0 (1):\penalty0 43--57, 2011.

\bibitem[Chatfield et~al.(2014{\natexlab{a}})Chatfield, Simonyan, Vedaldi, and
  Zisserman]{Chatfield14}
Ken Chatfield, Karen Simonyan, Andrea Vedaldi, and Andrew Zisserman.
\newblock Return of the devil in the details: Delving deep into convolutional
  nets.
\newblock In \emph{BMVC}, 2014{\natexlab{a}}.

\bibitem[Chatfield et~al.(2014{\natexlab{b}})Chatfield, Simonyan, and
  Zisserman]{ChatfieldOTF14}
Ken Chatfield, Karen Simonyan, and Andrew Zisserman.
\newblock Efficient on-the-fly category retrieval using convnets and {GPUs}.
\newblock In \emph{ACCV}, 2014{\natexlab{b}}.

\bibitem[Cheng et~al.(2015)Cheng, Yu, Feris, Kumar, Choudhary, and
  Chang]{CircProj}
Yu~Cheng, Felix~X. Yu, Rog{\'{e}}rio~Schmidt Feris, Sanjiv Kumar, Alok~N.
  Choudhary, and Shih{-}Fu Chang.
\newblock An exploration of parameter redundancy in deep networks with
  circulant projections.
\newblock In \emph{ICCV}, 2015.

\bibitem[Chetlur et~al.(2014)Chetlur, Woolley, Vandermersch, Cohen, Tran,
  Catanzaro, and Shelhamer]{cudnn}
Sharan Chetlur, Cliff Woolley, Philippe Vandermersch, Jonathan Cohen, John
  Tran, Bryan Catanzaro, and Evan Shelhamer.
\newblock cudnn: Efficient primitives for deep learning.
\newblock \emph{CoRR}, abs/1410.0759, 2014.

\bibitem[Collobert et~al.(2011)Collobert, Kavukcuoglu, and Farabet]{Torch}
Ronan Collobert, Koray Kavukcuoglu, and Cl{\'{e}}ment Farabet.
\newblock Torch7: A matlab-like environment for machine learning.
\newblock In \emph{BigLearn, NIPS Workshop}, 2011.

\bibitem[Deng et~al.(2009)Deng, Dong, Socher, Li, Li, and Li]{Imagenet}
Jia Deng, Wei Dong, Richard Socher, Li{-}Jia Li, Kai Li, and Fei{-}Fei Li.
\newblock Imagenet: {A} large-scale hierarchical image database.
\newblock In \emph{CVPR}, 2009.

\bibitem[Denton et~al.(2014)Denton, Zaremba, Bruna, LeCun, and
  Fergus]{Denton2014}
Emily~L. Denton, Wojciech Zaremba, Joan Bruna, Yann LeCun, and Rob Fergus.
\newblock Exploiting linear structure within convolutional networks for
  efficient evaluation.
\newblock In \emph{NIPS}, 2014.

\bibitem[Doll\'ar et~al.(2010)Doll\'ar, Welinder, and Perona]{Dollar2010}
Piotr Doll\'ar, Peter Welinder, and Pietro Perona.
\newblock Cascaded pose regression.
\newblock In \emph{CVPR}, 2010.

\bibitem[Girshick(2015)]{frcnn}
Ross~B. Girshick.
\newblock Fast {R-CNN}.
\newblock In \emph{ICCV}, 2015.

\bibitem[Han et~al.(2015)Han, Leung, Jia, Sukthankar, and Berg]{MatchNet15}
Xufeng Han, Thomas Leung, Yangqing Jia, Rahul Sukthankar, and Alexander~C.
  Berg.
\newblock Matchnet: Unifying feature and metric learning for patch-based
  matching.
\newblock In \emph{CVPR}, 2015.

\bibitem[He and Sun(2015)]{HeSun14}
Kaiming He and Jian Sun.
\newblock Convolutional neural networks at constrained time cost.
\newblock In \emph{CVPR}, 2015.

\bibitem[He et~al.(2016)He, Zhang, Ren, and Sun]{residuals}
Kaiming He, Xiangyu Zhang, Shaoqing Ren, and Jian Sun.
\newblock Deep residual learning for image recognition.
\newblock In \emph{CVPR}, 2016.

\bibitem[Hinton et~al.(2014)Hinton, Vinyals, and Dean]{Distilling}
Geoffrey~E. Hinton, Oriol Vinyals, and Jeffrey Dean.
\newblock Distilling the knowledge in a neural network.
\newblock \emph{Deep Learning and Representation Learning, NIPS Workshop},
  2014.

\bibitem[Ioannou et~al.(2016)Ioannou, Robertson, Cipolla, and
  Criminisi]{DeepRoots}
Yani Ioannou, Duncan~P. Robertson, Roberto Cipolla, and Antonio Criminisi.
\newblock Deep roots: Improving {CNN} efficiency with hierarchical filter
  groups.
\newblock \emph{CoRR}, abs/1605.06489, 2016.

\bibitem[Jaderberg et~al.(2014)Jaderberg, Vedaldi, and
  Zisserman]{Jaderberg2014}
Max Jaderberg, Andrea Vedaldi, and Andrew Zisserman.
\newblock Speeding up convolutional neural networks with low rank expansions.
\newblock In \emph{BMVC}, 2014.

\bibitem[J{\'{e}}gou et~al.(2012)J{\'{e}}gou, Perronnin, Douze, S{\'{a}}nchez,
  P{\'{e}}rez, and Schmid]{Jegou12}
Herv{\'{e}} J{\'{e}}gou, Florent Perronnin, Matthijs Douze, Jorge
  S{\'{a}}nchez, Patrick P{\'{e}}rez, and Cordelia Schmid.
\newblock Aggregating local image descriptors into compact codes.
\newblock \emph{IEEE Transactions on Pattern Analysis \& Machine Intelligence},
  34\penalty0 (9):\penalty0 1704--1716, 2012.

\bibitem[Kira et~al.(2012)Kira, Hadsell, Salgian, and Samarasekera]{LongRange}
Zsolt Kira, Raia Hadsell, Garbis Salgian, and Supun Samarasekera.
\newblock Long-range pedestrian detection using stereo and a cascade of
  convolutional network classifiers.
\newblock In \emph{IROS}, 2012.

\bibitem[Krizhevsky et~al.(2012)Krizhevsky, Sutskever, and
  Hinton]{Krizhevsky12}
Alex Krizhevsky, Ilya Sutskever, and Geoffrey~E. Hinton.
\newblock Imagenet classification with deep convolutional neural networks.
\newblock In \emph{NIPS}, 2012.

\bibitem[Lee et~al.(2015)Lee, Xie, Gallagher, Zhang, and Tu]{Lee14}
Chen{-}Yu Lee, Saining Xie, Patrick~W. Gallagher, Zhengyou Zhang, and Zhuowen
  Tu.
\newblock Deeply-supervised nets.
\newblock In \emph{AISTATS}, 2015.

\bibitem[Lenc et~al.(2011)Lenc, Gulshan, and Vedaldi]{lenc12vlbenchmarks}
K.~Lenc, V.~Gulshan, and A.~Vedaldi.
\newblock {VLBenchmarks}.
\newblock \url{http://www.vlfeat.org/benchmarks/}, 2011.

\bibitem[Li et~al.(2016)Li, Yosinski, Clune, Lipson, and
  Hopcroft]{ConvLearning}
Yixuan Li, Jason Yosinski, Jeff Clune, Hod Lipson, and John~E. Hopcroft.
\newblock Convergent learning: Do different neural networks learn the same
  representations?
\newblock In \emph{ICLR}, 2016.

\bibitem[Massa(2016)]{fmassaOD}
Francisco Massa.
\newblock Object detection in torch.
\newblock \url{http://github.com/fmassa/object-detection.torch}, 2016.

\bibitem[Mikolajczyk and Schmid(2005)]{Mikolajczyk}
Krystian Mikolajczyk and Cordelia Schmid.
\newblock A performance evaluation of local descriptors.
\newblock \emph{IEEE Transactions on Pattern Analysis \& Machine Intelligence},
  27\penalty0 (10):\penalty0 1615--1630, 2005.

\bibitem[Pinheiro et~al.(2015)Pinheiro, Collobert, and Doll{\'{a}}r]{DeepMask}
Pedro H.~O. Pinheiro, Ronan Collobert, and Piotr Doll{\'{a}}r.
\newblock Learning to segment object candidates.
\newblock In \emph{NIPS}, 2015.

\bibitem[Ren et~al.(2015)Ren, He, Girshick, and Sun]{faster}
Shaoqing Ren, Kaiming He, Ross~B. Girshick, and Jian Sun.
\newblock Faster {R-CNN:} towards real-time object detection with region
  proposal networks.
\newblock \emph{NIPS}, 2015.

\bibitem[Romero et~al.(2014)Romero, Ballas, Kahou, Chassang, Gatta, and
  Bengio]{FitNets}
Adriana Romero, Nicolas Ballas, Samira~Ebrahimi Kahou, Antoine Chassang, Carlo
  Gatta, and Yoshua Bengio.
\newblock Fitnets: Hints for thin deep nets.
\newblock \emph{CoRR}, abs/1412.6550, 2014.

\bibitem[Simo-Serra et~al.(2015)Simo-Serra, Trulls, Ferraz, Kokkinos, Fua, and
  Moreno-Noguer]{Serra15}
Edgar Simo-Serra, Eduard Trulls, Luis Ferraz, Iasonas Kokkinos, Pascal Fua, and
  Francesc Moreno-Noguer.
\newblock Discriminative learning of deep convolutional feature point
  descriptors.
\newblock In \emph{ICCV}, 2015.

\bibitem[Simonyan and Zisserman(2014)]{Vgg14}
Karen Simonyan and Andrew Zisserman.
\newblock Very deep convolutional networks for large-scale image recognition.
\newblock \emph{CoRR}, abs/1409.1556, 2014.

\bibitem[Sun et~al.(2013)Sun, Wang, and Tang]{Yi13}
Yi~Sun, Xiaogang Wang, and Xiaoou Tang.
\newblock Deep convolutional network cascade for facial point detection.
\newblock In \emph{CVPR}, 2013.

\bibitem[Toshev and Szegedy(2014)]{DeepPose}
Alexander Toshev and Christian Szegedy.
\newblock Deeppose: Human pose estimation via deep neural networks.
\newblock In \emph{CVPR}, 2014.

\bibitem[Viola and Jones(2001)]{ViolaJones}
Paul~A. Viola and Michael~J. Jones.
\newblock Rapid object detection using a boosted cascade of simple features.
\newblock In \emph{CVPR}, 2001.

\bibitem[Yan et~al.(2015)Yan, Zhang, Piramuthu, Jagadeesh, DeCoste, Di, and
  Yu]{hdcnn}
Zhicheng Yan, Hao Zhang, Robinson Piramuthu, Vignesh Jagadeesh, Dennis DeCoste,
  Wei Di, and Yizhou Yu.
\newblock Hd-cnn: Hierarchical deep convolutional neural network for large
  scale visual recognition.
\newblock In \emph{ICCV}, 2015.

\bibitem[Zagoruyko and Komodakis(2015)]{SZ15}
Sergey Zagoruyko and Nikos Komodakis.
\newblock Learning to compare image patches via convolutional neural networks.
\newblock In \emph{CVPR}, 2015.

\bibitem[{\v{Z}}bontar and LeCun(2015)]{Zbontar15}
Jure {\v{Z}}bontar and Yann LeCun.
\newblock Stereo matching by training a convolutional neural network to compare
  image patches.
\newblock \emph{CoRR}, abs/1510.05970, 2015.

\bibitem[Zehnder et~al.(2008)Zehnder, Koller{-}Meier, and Gool]{Zehnder}
Philipp Zehnder, Esther Koller{-}Meier, and Luc J.~Van Gool.
\newblock An efficient shared multi-class detection cascade.
\newblock In \emph{BMVC}, 2008.

\bibitem[Zhang et~al.(2015)Zhang, Zou, He, and Sun]{Zhang15}
Xiangyu Zhang, Jianhua Zou, Kaiming He, and Jian Sun.
\newblock Accelerating very deep convolutional networks for classification and
  detection.
\newblock \emph{CoRR}, abs/1505.06798, 2015.

\bibitem[Zheng et~al.(2015)Zheng, David, Georgescu, Nguyen, and
  Comaniciu]{Zheng15}
Yefeng Zheng, Liu David, Bogdan Georgescu, Hien Nguyen, and Dorin Comaniciu.
\newblock 3d deep learning for efficient and robust landmark detection in
  volumetric data.
\newblock In \emph{MICCAI}, 2015.

\end{thebibliography}
 
\end{document}